\documentclass[10pt,twocolumn]{article}

\newif\ifreview
\reviewfalse

\usepackage[utf8]{inputenc}
\usepackage[T1]{fontenc}
\usepackage{newtxtext}
\usepackage{newtxmath}
\usepackage{microtype}
\usepackage{inconsolata}
\usepackage{amsmath}
\usepackage{booktabs}
\usepackage{multirow}
\usepackage{graphicx}
\usepackage{xcolor}
\usepackage{hyperref}
\usepackage[capitalize,nameinlink]{cleveref}
\usepackage{caption}
\usepackage{enumitem}
\usepackage{listings}
\usepackage{balance}
\usepackage{url}
\usepackage{needspace}
\usepackage[letterpaper,left=0.72in,right=0.72in,top=1.02in,bottom=1.08in]{geometry}
\usepackage{titling}
\usepackage{titlesec}
\ifreview
\usepackage[mathlines,left]{lineno}
\fi
\usepackage{pgfplots}
\usepgfplotslibrary{groupplots}
\usetikzlibrary{patterns}
\pgfplotsset{compat=1.18}

\setlist[itemize]{leftmargin=*,topsep=2pt,itemsep=1pt}
\setlist[enumerate]{leftmargin=*,topsep=2pt,itemsep=1pt}
\titlespacing*{\section}{0pt}{8pt plus 2pt minus 1pt}{4pt plus 1pt minus 1pt}
\titlespacing*{\subsection}{0pt}{6pt plus 1pt minus 1pt}{3pt plus 1pt minus 1pt}
\newcommand{\paperpdfauthors}{Dongjie Chen, Ping Zhao, Bohua Zhan, Yulong Wang, Shushu Chen, Liangjun Feng, Hao Zhou, Min Shen, Linmu Wang, Weijia Sheng, Xiangyu Wei, Weijie Ding, Jianhui Huang, Yaoqing Gao}
\author{%
  \parbox{0.94\textwidth}{\centering\paperpdfauthors}\\[0.5em]
  Huawei Technologies Co., Ltd.
}

\hypersetup{
  colorlinks=true,
  linkcolor=blue,
  citecolor=blue,
  urlcolor=blue,
  pdfpagemode=UseOutlines,
  bookmarksopen=true,
  bookmarksopenlevel=1,
  bookmarksnumbered=true,
  pdftitle={Compiler-Grounded Hierarchical Diagnosis for LLM-Based Triton Kernel Optimization},
  pdfauthor={\paperpdfauthors}
}
\ifreview

\fi

\definecolor{paperindigo}{HTML}{3F5FB7}
\definecolor{papercoral}{HTML}{C76545}
\definecolor{paperteal}{HTML}{167C80}
\definecolor{paperslate}{HTML}{8A96A8}
\definecolor{paperink}{HTML}{27364A}
\definecolor{papergrid}{HTML}{DDE3EA}
\definecolor{codebg}{HTML}{F6F8FB}
\definecolor{codeframe}{HTML}{C9D3E0}
\definecolor{diffadd}{HTML}{216E68}
\definecolor{diffdel}{HTML}{B0543C}
\definecolor{diffmeta}{HTML}{697586}
\definecolor{codekw}{HTML}{3F5FB7}
\definecolor{codecomment}{HTML}{667085}
\definecolor{codeident}{HTML}{1F2937}
\definecolor{codestring}{HTML}{167C80}
\definecolor{codemeta}{HTML}{A85F3D}
\lstdefinelanguage{DiffPatch}{
  sensitive=true,
  morecomment=[l][\color{diffmeta}]{@@},
  morecomment=[l][\color{diffadd}]{+},
  morecomment=[l][\color{diffdel}]{-}
}
\lstdefinestyle{diffstyle}{
  basicstyle=\ttfamily\footnotesize,
  language=DiffPatch,
  xleftmargin=1pt,
  xrightmargin=1pt
}
\lstdefinestyle{runningexamplestyle}{
  language=Python,
  morekeywords={triton,jit},
  basicstyle=\ttfamily\footnotesize,
  keywordstyle=\color{codekw}\bfseries,
  commentstyle=\color{codecomment},
  stringstyle=\color{codestring},
  emph={small_row_kernel,medium_row_kernel,large_row_kernel,launch_by_rows},
  emphstyle=\color{codekw},
  emph={[2]BLOCK_ROWS},
  emphstyle={[2]\color{codemeta}\bfseries},
  xleftmargin=1pt,
  xrightmargin=1pt
}
\newenvironment{papercodeblock}
  {\par\addvspace{6pt}\noindent\begingroup\centering\begin{minipage}{0.94\columnwidth}}
  {\end{minipage}\par\endgroup\addvspace{6pt}}
\newcommand{\paperanalysisnote}[2]{%
  \Needspace{4\baselineskip}%
  \par\addvspace{3pt}%
  {\small\noindent\raggedright\emph{#1} #2\par}%
  \addvspace{3pt}}

\newcommand{\system}{the system}
\newcommand{\NPUTotalEntries}{37}
\newcommand{\NPUAnyValidEntries}{37}

\newcommand{\NPUAboveTwoXAnyValidEntries}{22}
\newcommand{\NPUAboveFiveXAnyValidEntries}{13}
\newcommand{\NPUTotalValidCases}{185}

\newcommand{\NPUGeomeanAnyValidSpeedup}{4.35}
\newcommand{\NPUMedianAnyValidSpeedup}{2.73}
\newcommand{\NPUKernelBenchSummaryTable}{%
\begin{table}[t]
\centering
\caption{Triton optimization speedup summary. Figure~\ref{fig:npukernelbench-runtime} reports the corresponding runtimes.}
\label{tab:npukernelbench-summary}
\small
\setlength{\tabcolsep}{4pt}
\begin{tabular}{lrrr}
\toprule
\textbf{Metric} & \textbf{Level 1} & \textbf{Level 2} & \textbf{Overall} \\
\midrule
\# Operators & 19 & 18 & 37 \\
Speedup (geometric mean) & 7.29$\times$ & 2.51$\times$ & 4.35$\times$ \\
Speedup (median) & 9.05$\times$ & 1.58$\times$ & 2.73$\times$ \\
\bottomrule
\end{tabular}
\end{table}
}
\newcommand{\NPUKernelBenchSpeedupFigure}{%
\begin{figure*}[t]
\centering
\begin{tikzpicture}
\begin{axis}[
    width=\linewidth, height=0.24\textwidth,
    ylabel={Triton optimization speedup},
    ymode=log,
    xmin=1, xmax=37, ymin=0.9000, ymax=80.17,
    xtick={1,10,20,30,37},
    ytick={0.1,0.2,0.5,1,2,5,10,20,50,100},
    yticklabels={$0.1\times$,$0.2\times$,$0.5\times$,$1\times$,$2\times$,$5\times$,$10\times$,$20\times$,$50\times$,$100\times$},
    yticklabel style={font=\scriptsize}, tick label style={font=\scriptsize},
    xlabel style={font=\scriptsize}, ylabel style={font=\scriptsize},
    xmajorgrids, x grid style={draw=papergrid},
    legend style={font=\scriptsize, draw=none, fill=white, fill opacity=0.9, text opacity=1, at={(0.5,0.98)}, anchor=north, legend columns=2},
]
\addplot[only marks, mark=*, mark size=1.75pt, paperindigo] coordinates {(4,1.0780) (6,1.2989) (7,1.3428) (14,1.9328) (15,1.9774) (16,2.0855) (17,2.2632) (20,3.4797) (24,4.4797) (26,9.0546) (27,10.3117) (29,12.0565) (30,13.7995) (31,17.4611) (32,17.8565) (34,43.5376) (35,44.3860) (36,66.0119) (37,76.3497)};
\addlegendentry{Level 1}
\addplot[only marks, mark=*, mark size=1.75pt, paperteal] coordinates {(1,1.0000) (2,1.0000) (3,1.0000) (5,1.2974) (8,1.3600) (9,1.3838) (10,1.4131) (11,1.5436) (12,1.5716) (13,1.5845) (18,2.7125) (19,2.7283) (21,4.1735) (22,4.1858) (23,4.2377) (25,7.8976) (28,11.8956) (33,23.6856)};
\addlegendentry{Level 2}
\addplot[densely dashed, paperink!40, line width=0.45pt, forget plot] coordinates {(1,1.0) (37,1.0)};
\addplot[densely dashed, paperink!40, line width=0.45pt, forget plot] coordinates {(1,2.0) (37,2.0)};
\addplot[densely dashed, paperink!40, line width=0.45pt, forget plot] coordinates {(1,5.0) (37,5.0)};
\addplot[densely dashed, paperink!24, line width=0.4pt, forget plot] coordinates {(1,10.0) (37,10.0)};
\addplot[densely dashed, paperink!24, line width=0.4pt, forget plot] coordinates {(1,20.0) (37,20.0)};
\addplot[densely dashed, paperink!24, line width=0.4pt, forget plot] coordinates {(1,50.0) (37,50.0)};
\node[anchor=north west, align=left, font=\scriptsize, fill=white, fill opacity=0.88, text opacity=1, rounded corners=1pt] at (rel axis cs:0.03,0.96) {Geo./median: 4.35$\times$ / 2.73$\times$\\$\geq$1.0/2.0/5.0$\times$: 37/37, 22/37, 13/37};
\end{axis}
\end{tikzpicture}
\caption{Distribution of Triton kernel optimization speedups, ordered by speedup.}
\label{fig:npukernelbench-speedup}
\end{figure*}
}
\newcommand{\NPUKernelBenchRuntimeFigure}{%
\begin{figure}[t]
\centering
\begin{tikzpicture}
\begin{axis}[
    ybar, bar width=6pt, width=\linewidth, height=0.46\columnwidth,
    enlarge y limits={upper, value=0.24},
    ylabel={Kernel time (geometric mean, $\mu$s)}, symbolic x coords={Level 1,Level 2,Overall}, xtick=data, enlarge x limits=0.22,
    tick label style={font=\scriptsize}, xticklabel style={font=\scriptsize, align=center, yshift=-1pt}, yticklabel style={font=\scriptsize}, ylabel style={font=\scriptsize},
    ymajorgrids, major grid style={draw=papergrid},
    legend style={font=\scriptsize, draw=none, fill=white, fill opacity=0.9, text opacity=1, at={(0.27,0.98)}, anchor=north west, legend columns=1},
    legend image code/.code={\draw[##1, draw=none] (0cm,-0.08cm) rectangle (0.18cm,0.10cm);},
]
\addplot[bar shift=-7pt, fill=paperslate!26, draw=paperslate!82!paperink, pattern=north east lines, pattern color=paperslate!72] coordinates {(Level 1,142.57) (Level 2,17.72) (Overall,51.70)};
\addlegendentry{initial}
\node[font=\tiny, text=paperslate!92!paperink, anchor=south, xshift=-7pt, yshift=2pt] at (axis cs:Level 1,142.57) {142.57};
\node[font=\tiny, text=paperslate!92!paperink, anchor=south, xshift=-7pt, yshift=2pt] at (axis cs:Level 2,17.72) {17.72};
\node[font=\tiny, text=paperslate!92!paperink, anchor=south, xshift=-7pt, yshift=2pt] at (axis cs:Overall,51.70) {51.70};
\addplot[bar shift=7pt, fill=paperteal!82, draw=paperteal!88!paperink, nodes near coords, every node near coord/.append style={font=\tiny, text=paperteal!92!paperink, anchor=south, yshift=2pt}] coordinates {(Level 1,19.54) (Level 2,7.05) (Overall,11.90)};
\addlegendentry{optimized}
\node[anchor=north east, align=right, font=\scriptsize, fill=white, fill opacity=0.9, text opacity=1, rounded corners=1pt] at (rel axis cs:0.98,0.98) {Speedup L1/L2/All:\\7.29$\times$ / 2.51$\times$ / 4.35$\times$};
\end{axis}
\end{tikzpicture}
\caption{Geometric-mean Triton kernel time before and after optimization.}
\label{fig:npukernelbench-runtime}
\end{figure}
}
\newcommand{\NPUTorchTotalEntries}{35}
\newcommand{\NPUTorchTotalValidCases}{1678}
\newcommand{\NPUTorchMinimumCasesPerEntry}{14}
\newcommand{\NPUTorchMedianCasesPerEntry}{50}
\newcommand{\NPUTorchMaximumCasesPerEntry}{111}
\newcommand{\NPUTorchGeomeanSpeedup}{2.60}
\newcommand{\NPUTorchMedianSpeedup}{1.46}
\newcommand{\NPUTorchImprovingEntries}{22}
\newcommand{\NPUTorchAboveTwoXEntries}{16}
\newcommand{\NPUTorchAboveFiveXEntries}{14}

\newcommand{\NPUTorchComparisonTable}{%
\begin{table}[t]
\centering
\caption{Post-optimization comparison with Torch NPU. Speedup is Torch NPU time divided by optimized Triton total operator time.}
\label{tab:torch-npu-comparison}
\small
\setlength{\tabcolsep}{4pt}
\begin{tabular}{lrrrr}
\toprule
\textbf{Group} & \textbf{Operators} & \textbf{Cases} & \textbf{Geo. speedup} & \textbf{Median} \\
\midrule
Level 1 & 18 & 894 & 1.41$\times$ & 0.87$\times$ \\
Level 2 & 17 & 784 & 4.96$\times$ & 5.98$\times$ \\
Overall & 35 & 1678 & 2.60$\times$ & 1.46$\times$ \\
\bottomrule
\end{tabular}
\end{table}
}

\newcommand{\NPURoundAccountingEntries}{37}

\newcommand{\NPUBestRoundThirteenToFifteenCount}{12}
\newcommand{\NPUMedianBestRound}{10}
\newcommand{\NPUBestRoundNineToFifteenCount}{22}
\newcommand{\NPUWinningLOneCount}{33}
\newcommand{\NPUWinningLTwoCount}{4}

\newcommand{\NPUPathUsesLTwoCount}{32}
\newcommand{\NPUPathUsesLThreeCount}{3}
\newcommand{\NPUPathUsesLFourCount}{4}

\newcommand{\NPUOptimizationAccountingFigure}{%
\begin{figure*}[t]
\centering
\begin{minipage}[t]{0.32\textwidth}
\centering
\begin{tikzpicture}[baseline=(current bounding box.north)]
\begin{axis}[
    ybar, bar width=8pt, width=\linewidth, height=0.62\linewidth,
    ymin=0, ymax=18, title={(a) Best validated round}, ylabel={Operators},
    symbolic x coords={1--4,5--8,9--12,13--15}, xtick=data,
    ymajorgrids, major grid style={draw=papergrid},
    tick label style={font=\scriptsize}, xticklabel style={font=\scriptsize, align=center},
    yticklabel style={font=\scriptsize}, title style={font=\scriptsize}, ylabel style={font=\scriptsize},
    nodes near coords, every node near coord/.append style={font=\scriptsize, fill=white, inner sep=0.5pt},
]
\addplot[fill=paperindigo!82, draw=paperindigo!88!paperink] coordinates {(1--4,11) (5--8,4) (9--12,10) (13--15,12)};
\end{axis}
\end{tikzpicture}
\end{minipage}\hfill
\begin{minipage}[t]{0.32\textwidth}
\centering
\begin{tikzpicture}[baseline=(current bounding box.north)]
\begin{axis}[
    ybar, bar width=8pt, width=\linewidth, height=0.62\linewidth,
    ymin=0, ymax=39, title={(b) Winning level},
    symbolic x coords={triage,profiling}, xtick=data,
    xticklabels={Pattern\\ triage,Profiling\\ diagnosis},
    enlarge x limits=0.36,
    ymajorgrids, major grid style={draw=papergrid},
    tick label style={font=\scriptsize}, xticklabel style={font=\scriptsize, align=center, text width=1.7cm},
    yticklabel style={font=\scriptsize}, title style={font=\scriptsize},
    nodes near coords, every node near coord/.append style={font=\scriptsize, fill=white, inner sep=0.5pt},
]
\addplot[fill=paperteal!82, draw=paperteal!88!paperink] coordinates {(triage,33) (profiling,4)};
\node[anchor=north east, align=right, font=\tiny, text=paperink] at (rel axis cs:0.97,0.94) {IR and compiler-source:\\0 selected best rounds};
\end{axis}
\end{tikzpicture}
\end{minipage}\hfill
\begin{minipage}[t]{0.32\textwidth}
\centering
\begin{tikzpicture}[baseline=(current bounding box.north)]
\begin{axis}[
    ybar, bar width=8pt, width=\linewidth, height=0.62\linewidth,
    ymin=0, ymax=43, title={(c) Best path reaches level},
    symbolic x coords={triage,profiling,attribution,compiler}, xtick=data,
    xticklabels={Pattern\\ triage,Profiling\\ diagnosis,IR\\ attribution,Compiler-\\ source\\ escalation},
    enlarge x limits=0.20,
    ymajorgrids, major grid style={draw=papergrid},
    tick label style={font=\scriptsize}, xticklabel style={font=\tiny, align=center, text width=1.55cm},
    yticklabel style={font=\scriptsize}, title style={font=\scriptsize},
    nodes near coords, every node near coord/.append style={font=\scriptsize, fill=white, inner sep=0.5pt},
]
\addplot[fill=papercoral!82, draw=papercoral!88!paperink] coordinates {(triage,37) (profiling,32) (attribution,3) (compiler,4)};
\end{axis}
\end{tikzpicture}
\end{minipage}
\caption{Optimization accounting. Panel (a) records the round with the best validated, comparable per-round result; panel (b) records the primary evidence level of that selected round; panel (c) records whether the complete trajectory reaches each level at least once. Thus, panel (b) does not measure whether deeper analysis was used before the selected round. The median best-round index is 10; 28/37 winning-round levels are explicitly labeled by the optimization process, with the remainder classified from their associated evidence.}
\label{fig:npu-optimization-accounting}
\end{figure*}
}

\title{Compiler-Grounded Hierarchical Diagnosis for LLM-Based Triton Kernel Optimization}
\date{}

\begin{document}
\maketitle
\ifreview
\modulolinenumbers[5]
\linenumbers
\fi

\begin{abstract}
Recent advances in large language models (LLMs) have enabled automated kernel generation and optimization, but most existing approaches rely on surface signals such as compilation feedback and profiling metrics.
These signals reveal that a kernel is slow, but not why the backend compiler fails to realize a profitable optimization, especially on emerging accelerators such as NPUs.
We therefore formulate kernel optimization as a progressive cross-layer diagnosis problem that links runtime symptoms to IR structure and compiler behavior before rewriting source.
Based on this insight, we present our system, a compiler-grounded and hierarchical optimization framework for Triton kernels.
\system{} escalates from lightweight pattern triage and profiling diagnosis to IR attribution and compiler-grounded analysis only when deeper evidence is needed, then proposes evidence-backed source-level rewrites.

We implement \system{} on Triton for Ascend NPUs and evaluate it on \NPUTotalEntries{} successfully converted entries from a standardized NPUKernelBench-derived Ascend 950 benchmark.
Across these entries, \system{} attains a geometric-mean speedup of \NPUGeomeanAnyValidSpeedup$\times$ and a median speedup of \NPUMedianAnyValidSpeedup$\times$ from the initial to optimized Triton kernel; \NPUAboveTwoXAnyValidEntries/\NPUAnyValidEntries{} exceed 2$\times$ and \NPUAboveFiveXAnyValidEntries/\NPUAnyValidEntries{} exceed 5$\times$.
The complete distribution ranges from near-baseline entries to large wins, motivating transparent reporting of the current system's scope and limitations.
\end{abstract}

\section{Introduction}

Efficient kernel implementation is critical for achieving high performance on modern accelerators.
Domain-specific languages such as Triton have significantly lowered the barrier for writing high-performance kernels~\cite{tillet2019triton}, enabling developers to express optimized computations with relatively concise code.
Meanwhile, recent advances in large language models (LLMs) have inspired a new line of work on LLM-driven kernel generation and optimization, where code agents iteratively generate, execute, and refine kernels based on runtime feedback.
Recent surveys and curated repositories show that this area is expanding quickly across benchmarks, agentic optimization systems, and accelerator-specific kernel generation pipelines~\cite{yu2026automatedkernelgeneration,flagos2026awesome}.

Most existing LLM-based approaches, including recent profile-guided optimization systems, still optimize primarily at the source-code level~\cite{li2025tritonforge,lei2025pragma,wei2025astra}.
They rely on shallow signals such as compilation errors or runtime metrics, which can reveal that a kernel is slow but rarely explain why it is slow.
In practice, many performance failures emerge only after the kernel is lowered through intermediate representation (IR) stages, scheduling decisions, and backend-specific compiler passes.
As a result, there is a fundamental semantic gap between operator-level code and backend performance behavior.

This gap is particularly severe on emerging accelerators such as NPUs (e.g., Ascend), where the Triton-Ascend stack is still evolving and backend behavior is target-specific and relatively opaque to developers~\cite{tritonascendrepo}.
LLM agents, which are typically pretrained on GPU-centric code patterns, therefore behave like trial-and-error optimizers: they propose plausible source edits, but they lack the evidence needed to map a runtime symptom to a compiler-level cause.

The key insight of this paper is that kernel optimization should be treated as a \emph{progressive cross-layer diagnosis problem}.
A useful optimizer should first exploit low-cost evidence, then escalate to deeper compiler grounding only when shallower evidence is insufficient.
In this view, compiler grounding is not an isolated feature; it is the deepest layer in a hierarchical reasoning process that connects
\begin{quote}
\centering
\emph{profiling signals} $\rightarrow$ \emph{IR-level evidence} $\rightarrow$ \emph{compiler transformations} $\rightarrow$ \emph{source-level rewrites}.
\end{quote}

Based on this insight, we propose a compiler-grounded optimization system for Triton kernels.
\system{} forms a closed-loop optimization pipeline that starts from pattern triage, escalates through profiling diagnosis and IR attribution, invokes compiler-source escalation when needed, and finally proposes evidence-backed source-level rewrites.
The hierarchy makes compiler grounding practical: deeper compiler-side reasoning is invoked only when lighter-weight evidence cannot yet produce a sufficiently grounded rewrite.
Beyond this per-operator online loop, the current system also performs a slower retrospective synthesis pass over validated benchmark batches so that repeated optimization histories can refine existing pattern guidance, reduce repeated diagnostic cost across operators, and introduce new reusable pattern families for later runs.

We implement \system{} on top of Triton for Ascend NPUs and evaluate it with an OpenCode agent backed by DeepSeek V4 Pro on a NPUKernelBench-derived Ascend 950 suite.
After removing failed Torch-to-Triton conversions, the evaluation cohort contains \NPUTotalEntries{} successfully converted operators.
For each operator, five cases sampled from its original benchmark inputs form the optimization benchmark, for \NPUTotalValidCases{} cases in total.
Across these entries, the geometric-mean initial-to-optimized Triton kernel speedup is \NPUGeomeanAnyValidSpeedup$\times$ and the median is \NPUMedianAnyValidSpeedup$\times$.
The distribution is deliberately reported in full: \NPUAboveTwoXAnyValidEntries/\NPUAnyValidEntries{} entries exceed 2$\times$ and \NPUAboveFiveXAnyValidEntries/\NPUAnyValidEntries{} exceed 5$\times$.
Thus, the results demonstrate substantial headroom on a subset of operators but not uniform improvement.
The per-round optimization trajectories further show that \NPUBestRoundNineToFifteenCount{} entries first reach their best validated result after round~8, while \NPUPathUsesLTwoCount{} selected paths reach profiling diagnosis, \NPUPathUsesLThreeCount{} reach IR attribution, and \NPUPathUsesLFourCount{} reach compiler-source escalation.
We therefore present the paper as a systems argument with transparent benchmark and process evidence, reserving causal decomposition of the hierarchy for future controlled experiments.

In summary, this paper makes the following contributions:

\begin{itemize}
    \item We formulate LLM-based kernel optimization as a progressive cross-layer diagnosis problem and identify the lack of compiler-grounded evidence as a key reason profiling-only agents fail on backend-specific optimizations.
    \item We describe a hierarchical workflow that escalates from pattern triage to profiling diagnosis, IR attribution, and compiler-source escalation before proposing evidence-backed source-level rewrites.
    \item We present standardized, distribution-aware evaluation evidence on a NPUKernelBench-derived Ascend 950 benchmark, reporting aggregate gains, best-round and evidence-depth accounting, and a representative cross-layer trace.
\end{itemize}

\section{Motivation}

\subsection{A Running Example}

Consider a Mixture-of-Experts group-score Triton kernel optimized in our experiment.
For every group of 32 expert scores, the kernel must compute the sum of its two largest values.
Its initial implementation obtains the first maximum, uses \texttt{tl.argmax} to identify the corresponding index, masks that index, and then reduces again for the second maximum.
\Cref{fig:motivation-moe} shows the later source rewrite selected after cross-layer analysis.

\begin{figure}[t]
\centering
\begin{papercodeblock}
\begin{lstlisting}[style=diffstyle]
 group_values = tl.gather(all_scores, group_offsets, 0)
-top1 = tl.max(group_values, axis=0)
-top1_idx = tl.argmax(group_values, axis=0)
-rest = tl.where(offsets == top1_idx, -float("inf"), group_values)
-group_score = top1 + tl.max(rest, axis=0)
+pair_sums = group_values[:, None] + group_values[None, :]
+off_diagonal = offsets[:, None] != offsets[None, :]
+pair_sums = tl.where(off_diagonal, pair_sums, -float("inf"))
+group_score = tl.max(tl.max(pair_sums, axis=1), axis=0)
\end{lstlisting}
\end{papercodeblock}
\caption{Actual source rewrite for the MoE group-score kernel.  The maximum off-diagonal pair sum is mathematically equal to the sum of the two largest group values, so the rewrite removes index-producing reduction while preserving ties.}
\label{fig:motivation-moe}
\end{figure}

Profiling identifies a mixed execution profile: 71\% vector activity but 26\% scalar activity.
This leaves several plausible source-level reactions, such as forcing a scalar-oriented SIMT mode or replacing \texttt{argmax} with value-based masking.
Neither is sufficient: forced SIMT severely regresses because it disables the productive vector path, while value-based masking fails correctness on low-precision ties.
Profiling therefore locates a symptom but does not identify a tie-correct way to remove it.

IR attribution then identifies 26 synchronization operations in the early lowered stages and ties them to the \texttt{argmax} reductions and index-mask construction.
Compiler-source analysis explains the mechanism: vector maxima are supported natively, whereas \texttt{argmax} requires an additional scalar index-tracking stage with synchronization.
This evidence rules out launch-mode tuning as the remedy and constrains a successful rewrite to avoid index-producing reduction altogether.

The resulting algebraic reformulation is shown in \Cref{fig:motivation-moe}: $\max_{i \ne j}(v_i + v_j)$ is exactly the sum of the two largest group values, including tied maxima.
It replaces the \texttt{argmax}-and-mask chain with vectorized pairwise addition and value-only reduction, and passes all 50 correctness cases.
The optimized kernel reaches 1.42$\times$ geometric-mean speedup.

This is the failure mode we care about.
A profile can identify a scalar-heavy path, but it cannot determine whether scalar work originates in source control flow, an IR-level lowering, or a backend implementation requirement.
The surface symptom is visible, but the lowering mechanism that distinguishes these hypotheses is hidden behind compiler transformations.
That gap is exactly where source-level trial-and-error becomes unreliable.

This motivates the central challenge of the paper:
\emph{the relationship between source-level code and runtime performance is mediated by compiler transformations that are invisible to surface-level optimization signals.}

\subsection{Limitations of Existing Approaches}

Existing LLM-based kernel systems have made rapid progress, but most of them still optimize at the level of source generation or source mutation.
Source-level reward-loop methods such as SparseRL~\cite{wang2026sparserl}, Kevin~\cite{baronio2025kevin}, CUDA-L1~\cite{li2025cudal1}, and KernelEvolve~\cite{liao2025kernelevolve} demonstrate that runtime reward can substantially improve correctness and throughput.
Their core abstraction, however, remains the same: kernel text is the object of search, while the compiler is treated as a black box that returns either performance or failure signals.
That abstraction is powerful when the main challenge is broad exploration in source space, but it becomes brittle when multiple source rewrites induce similar runtime symptoms while diverging only after lowering.
In the running example above, changing execution mode or masking values by equality may both look reasonable under end-to-end reward, yet neither supplies the tie-correct, vectorized reformulation implied by the lowering evidence.
Without an explicit bridge from runtime symptoms to IR structure and pass behavior, the optimizer can only infer the cause indirectly from success or failure after execution.

Profiling-guided systems such as TritonForge, PRAGMA, Astra, and KForge move one step closer to diagnosis by adding hotspots, hardware-counter summaries, and multi-agent reasoning around execution traces~\cite{li2025tritonforge,lei2025pragma,wei2025astra,sereda2025kforge}.
This is a meaningful advance because it helps the optimizer prioritize where to intervene and reduces obviously unproductive rewrites.
However, profiling still localizes the symptom more reliably than the mechanism: it can show that a kernel is movement-heavy, scalar-heavy, or serialization-heavy, but not whether the root cause is work distribution, an IR-level layout artifact, or a backend pass precondition.
This limitation is especially important for NPU-oriented generation systems such as AscendKernelGen and AscendCraft~\cite{cao2026ascendkernelgen,wen2026ascendcraft}, where backend behavior is even more target-specific.
What is still missing is a progressive diagnosis path that carries the evidence from runtime behavior to IR structure and then to compiler constraints before choosing the rewrite.

\subsection{Key Insight}

From this observation, we derive the key insight of this work:

\begin{quote}
\centering
\emph{Effective kernel optimization should be cast as progressive diagnosis across layers, with deeper compiler grounding invoked only when shallower evidence is insufficient.}
\end{quote}

This is why compiler grounding should be selective rather than always-on: it is the mechanism that turns a vague slowdown into a rewrite that is both justified and backend-aware.

\section{System Overview}

\begin{figure*}[t]
\centering
\includegraphics[width=0.94\textwidth,keepaspectratio]{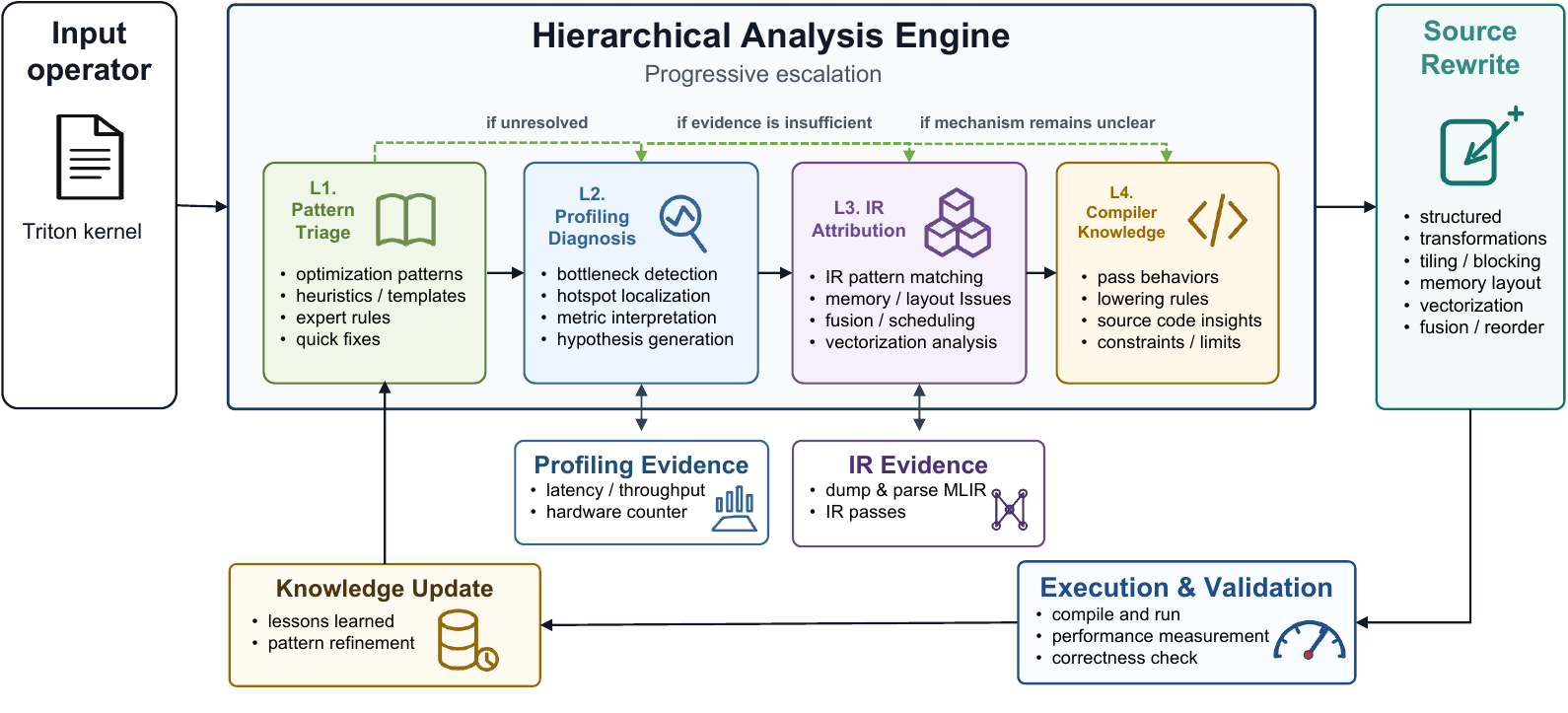}
\caption{Overview of \system. The loop escalates from pattern triage to profiling, IR attribution, and compiler-grounded analysis only when lower-cost evidence is insufficient, then proposes evidence-backed source-level rewrites.}
\label{fig:overview}
\end{figure*}

\system{} combines pattern triage, profiling diagnosis, IR attribution, and compiler-grounded reasoning in a unified loop.
At a high level, the system first asks whether a strong pattern-backed hypothesis already exists, then gathers runtime evidence from profiling, analyzes intermediate representations when needed to identify structural inefficiencies, and finally consults compiler-grounded references when shallower analysis is insufficient.
The output is an evidence-backed source-level rewrite that can be evaluated and iteratively refined.

This online optimization loop is paired with a slower cross-operator knowledge-synthesis loop.
After a batch benchmark sweep completes, \system{} rereads validated optimization histories across operators, aligns each round with an existing pattern family when possible, and promotes repeated new mechanisms into reusable guidance when no current family explains them well.
Future optimization sessions therefore start from a refreshed low-cost hypothesis layer rather than from a fixed hand-authored pattern seed.

The loop is organized around explicit optimization states rather than hidden conversational context.
Before the first code-changing round, \system{} establishes a validated baseline state that fixes the executable kernel, evaluation context, and reference metrics.
Each later round inherits that state, records its rewrite hypothesis and supporting evidence, and completes with a validated performance comparison that defines the next decision point.
Deeper profiling, IR, and compiler-grounded analyses therefore become reusable evidence attached to the round rather than ephemeral observations.
This stateful design makes the optimization process explicit, resumable, and auditable across rounds.

The optimization process can be run either as a single end-to-end session or as a sequence of externally audited rounds.
In both cases, the hierarchy in Figure~\ref{fig:overview} is enforced through explicit round boundaries, round-local evidence, and validation gates between optimization steps.

A key property of \system{} is that it does not rely on compiler-level reasoning for every optimization opportunity.
Instead, it adopts a hierarchical strategy in which inexpensive pattern triage is applied first, followed by progressively deeper reasoning only when required.
This design improves efficiency, reduces unnecessary context overload for the agent, and preserves interpretability throughout the optimization process.
In this sense, compiler grounding is not an isolated module, but the deepest diagnostic stage in a progressive cross-layer optimization loop whose cheapest stage can itself improve over time through retrospective synthesis.

\section{Method}

\system{} is designed around a central systems question: how can an optimization system access enough compiler information to diagnose backend-specific failures without forcing deep compiler reasoning on every optimization round?
\system{} answers this question with a hierarchical diagnostic workflow.
Each level consumes the evidence accumulated so far, either proposes a grounded rewrite hypothesis or escalates the optimization request to a deeper analysis stage.
This section describes the escalation workflow, the profiling-to-IR bridge, and compiler-grounded analysis.

\subsection{Hierarchical Diagnostic Workflow}

\system{} organizes optimization as four named levels of progressively more specific evidence.
In the current prototype, the optimization workflow traverses these levels as an ordered analysis ladder: \emph{pattern triage}, \emph{profiling diagnosis}, \emph{IR attribution}, and \emph{compiler-source escalation}.

At \textbf{L1} (\textbf{Pattern Triage}), the system asks whether a strong pattern-backed hypothesis already exists.
Operationally, the workflow first consults a curated pattern index and then opens only the one or two detailed pattern references that match the current symptom.
The corresponding pattern guidance mixes publicly documented Triton idioms, including autotuning, program reordering, block pointers, and compiler hints~\cite{tillet2019triton,tritondocs-matmul,tritondocs-autotune,tritondocs-language}, selected Ascend-oriented tuned operators and tutorial cases curated in prior public optimization materials~\cite{tritonascendopsrepo}, and additional internal patterns and anti-signals distilled from prior validated optimization histories.
Not every pattern therefore maps cleanly to a single external citation: some are adapted from curated examples in external repositories, while others are internal reusable heuristics extracted from prior rounds rather than externally sourced claims.
If code inspection, benchmark shape, and existing validated evidence already justify a low-risk rewrite, the round may stop here without invoking profiling or IR capture, but that decision must still be recorded explicitly in the round rationale.

L1 is therefore not a fixed knowledge source.
After batched benchmark optimization runs finish, \system{} performs a retrospective synthesis pass over the per-round optimization trajectories.
That pass semantically maps each round to an existing pattern family when possible, records the preconditions, rewrite, observed evidence, and interpretation needed to explain the outcome, and creates a new reusable pattern family when no current card matches the mechanism.
The result is a refreshed quick-match index for future sessions: pattern triage starts from accumulated cross-operator evidence rather than from isolated one-off successes.

At \textbf{L2} (\textbf{Profiling Diagnosis}), the system gathers runtime evidence, including latency, throughput, and hardware-counter summaries, to localize the bottleneck class.
At this stage, \system{} builds a structured execution view of the operator on the NPU, including kernel time, data-movement behavior, API-side overhead, and hardware utilization signals such as AIV/AIC occupancy and overlap.
This stage does not yet explain the compiler mechanism, but it determines whether the issue is likely dominated by memory behavior, pipeline overlap, or scheduling overhead.

At \textbf{L3} (\textbf{IR Attribution}), \system{} inspects intermediate representations (IR) to translate the runtime symptom into a structural diagnosis, such as non-unit-stride loads, missed fusion, or layout-induced vectorization failures.
Rather than treating lowering as an opaque transition, \system{} reconstructs the operator's stage-by-stage IR trajectory and organizes it as an evidence view that highlights stage summaries, adjacent-pass changes, and performance-relevant structural signatures.
If the IR diagnosis already maps cleanly to a known rewrite template, the system can act at this stage.

At \textbf{L4} (\textbf{Compiler-Source Escalation}), \system{} invokes compiler-grounded analysis to explain why the diagnosed pattern blocks backend optimization.
This stage asks a narrowed compiler question: which pass-level constraint, legality condition, or backend assumption prevents the candidate optimization from materializing?
Its purpose is to recover the source-level conditions that a successful rewrite must satisfy, rather than to conduct unconstrained compiler exploration.

The output of each level is therefore not just a candidate edit, but an increasingly specific diagnosis.
\system{} escalates only when the current level cannot produce a sufficiently grounded rewrite hypothesis.
Operationally, the policy uses three concrete tests.
First, the current level must identify a dominant hotspot or bottleneck class for the benchmark cases that still matter; if the evidence only says ``performance is still low'' without localizing the active cost, the round cannot stop.
Second, the current level must narrow the next change to one evidence-backed rewrite family; if two or more competing explanations remain plausible, the system escalates rather than choosing arbitrarily.
Third, the current level must justify the rewrite from evidence already exposed at that level; if the remaining question is about lowering behavior, tail handling, legality conditions, or pass-specific constraints, the policy escalates to IR attribution or compiler-source escalation instead of treating the effect as a source-level tuning problem.
Conversely, the workflow stops escalating once a stage can state a concrete rewrite hypothesis together with the evidence needed to justify why that rewrite should help on the current hotspot.
This escalation logic is recorded in the optimization history itself: every completed round records its hypothesis, evidence sources, comparison target, and next recommendation, and each baseline or candidate round passes validation before the session continues.
If a deeper stage still yields no actionable diagnosis, the round is recorded as inconclusive rather than forcing a speculative rewrite.
As a result, skipped evidence is visible and justified, failed escalations remain auditable, and deeper analysis leaves persistent diagnostic records that later rounds can reuse locally and that later batch synthesis can reuse globally.

\subsection{Profiling-to-IR Mapping}

Given a candidate kernel, \system{} first profiles the benchmarked operator under a standardized measurement protocol.
The profiling stage converts the raw trace into operator-level timing summaries and structured execution signals such as data-transfer pressure, API overhead, and device-side utilization or overlap indicators.
This gives the system a concrete hotspot target and a first classification of the likely bottleneck.

When timing evidence alone is not sufficient, \system{} reconstructs the same operator's lowering trajectory across Triton and downstream compiler IR.
The resulting provenance record ties the observed hotspot to specific stage-to-stage transformations, which makes the profiling-to-IR bridge explicit rather than implicit.

The resulting IR view summarizes stage-local signatures, pass-to-pass changes, and performance-relevant structural patterns.
This does not make profiler-to-IR attribution fully automatic, but it narrows the diagnosis to a concrete lowering path that can support the next rewrite decision.

\subsection{Compiler-Grounded Analysis}

Rather than treating the compiler as a black box, \system{} is designed to expose compiler-grounded evidence to the optimization workflow when benchmark, profiler, and IR evidence are no longer sufficient.
At this stage, \system{} first consults compiler knowledge that has already been distilled from earlier source analysis into pass-oriented references, narrowed question templates, and backend-specific invariants.
For example, one such reference may record that Ascend vectorized memory paths usually require aligned power-of-2 block widths together with explicit contiguity or alignment information, so a non-power-of-2 tile or missing contiguity hint can explain why an apparently contiguous load sequence remains scalarized.
These references let the system reformulate an IR symptom into a concrete compiler question, such as which pass family would explain a suspicious stage transition or which backend constraint would cause a vectorization loss or copy-growth pattern.
Only when those references still leave the answer unresolved does \system{} escalate to a focused source inspection of the specific compiler subsystem implicated by the current evidence.
The output is therefore not a general compiler explanation, but a localized account of why the current IR pattern blocks backend optimization and what source-level conditions a successful rewrite must satisfy.

This design keeps compiler grounding narrow and auditable.
Compiler reasoning is triggered only after earlier evidence has localized the question, so in practice it acts as a late explanation-and-constraint recovery stage rather than as another unconstrained search space.

\section{Implementation}

We implement \system{} as an agent-centric optimization environment built on top of existing code-agent tooling rather than as a bespoke compiler service.
The implementation described in this paper has been integrated into the CannBot skills repository.\footnote{\url{https://gitcode.com/cann/cannbot-skills}}
The implementation is organized around two layers: a thin execution harness, in the sense of harness engineering for code agents~\cite{openai-harness-engineering}, and a set of optimization skills aligned with the hierarchy.
This split is deliberate.
The harness is responsible for making evidence collection and validation reliable, while the skills carry the domain-specific optimization logic.
In this framing, the main engineering effort is not to hand-script each optimization decision, but to design the environment, specify the round contract, and build the feedback loops that allow a general-purpose code agent to perform reliable optimization work.

The harness establishes the validated baseline, runs correctness tests and benchmark comparisons, gathers profiler outputs, captures IR artifacts, and records every round as an explicit state with its parent, evidence, validation outcome, and performance comparison.
An explicit round-state machine enforces the lifecycle from a validated baseline through a proposed round, any evidence-driven change of analysis policy, and a submitted validation result.
At these boundaries, an agent hook checks the round contract---for example, that the intended evidence depth and the resulting validation record are present and mutually consistent---before the workflow advances.
This separates workflow compliance from optimization strategy: the state machine and hook make omissions and invalid transitions visible, while the agent remains responsible for selecting the next rewrite.
Its role is therefore infrastructural rather than strategic.
It provides the agent with stable artifacts, comparable measurements, and guardrails that reject invalid or regressing candidates early.
Because this layer is thin, the same protocol can be reused with different code agents, including Codex~\cite{openai-codex} and OpenCode~\cite{opencode-agent}.

The optimization logic itself is realized as a set of agent skills that mirror the hierarchy in Figure~\ref{fig:overview}.
Pattern-triage skills consult the curated pattern index and synthesize low-cost rewrite hypotheses from reference-oriented guidance.
Profiling skills convert execution traces into bottleneck diagnoses.
IR-attribution skills reconstruct the lowering trajectory and surface performance-relevant structural changes.
Compiler-grounded skills turn unresolved IR symptoms into narrow compiler questions and recover the source-level constraints that a successful rewrite must satisfy.
Finally, rewrite skills map these diagnoses into source-level edits that are validated against the current round's benchmark and correctness criteria.

This organization keeps the implementation close to the paper's core claim.
The harness enforces auditable optimization rounds, while the skills decide how to spend analytical depth within each round.
A retrospective synthesis pass then feeds validated mechanisms back into pattern guidance, so the cheapest level can improve over time without entangling reusable knowledge with the execution substrate.

\section{Evaluation}

We evaluate \system{} on an NPUKernelBench-derived suite that ports selected KernelBench operators to Ascend 950~\cite{ouyang2025kernelbench,npukernelbench2026}.
Our evaluation separates final operator-level outcomes, their distribution, and the evidence trajectory behind representative optimization decisions.

We organize the evaluation around three questions:

\begin{itemize}
    \item \textbf{RQ1:} What kernel-level speedups does \system{} obtain from the initial to optimized Triton implementation?
    \item \textbf{RQ2:} Are the outcomes broadly distributed, or concentrated in a small subset of operators?
    \item \textbf{RQ3:} Do per-round optimization trajectories show the intended progression from pattern triage to deeper evidence?
\end{itemize}

Table~\ref{tab:npukernelbench-summary} and Figure~\ref{fig:npukernelbench-runtime} answer RQ1; Figure~\ref{fig:npukernelbench-speedup} answers RQ2; and the optimization accounting plus the representative trace answer RQ3.
We separately report a post-optimization comparison against Torch NPU; it is not used as the before/after optimization metric.

\subsection{Experimental Setup}

NPUKernelBench is derived from KernelBench by porting selected operators to the Ascend NPU platform~\cite{ouyang2025kernelbench,npukernelbench2026}.
For each operator, the workflow begins with a PyTorch reference and a Triton implementation whose baseline is validated before optimization.
The optimization agent is OpenCode~\cite{opencode-agent} backed by DeepSeek V4 Pro~\cite{deepseekai2026deepseekv4}, targeting the A5 configuration of an Ascend 950 accelerator.
We optimize each operator for 15 rounds.

We remove operators whose AI conversion from the Torch NPU implementation fails, leaving \NPUTotalEntries{} successfully converted Triton operators.
For each retained operator, we sample five cases from its original benchmark inputs to form the fixed benchmark used throughout optimization, giving \NPUTotalValidCases{} kernel benchmark cases in total.
We take the geometric mean over the five cases for each operator; all headline values are then geometric means across operators.
The main result is thus a kernel-level initial-to-optimized Triton comparison, not an end-to-end application latency result.

\NPUKernelBenchSummaryTable
\NPUKernelBenchSpeedupFigure
\NPUKernelBenchRuntimeFigure
\NPUOptimizationAccountingFigure

\subsection{RQ1: Aggregate Effectiveness}

Across the \NPUAnyValidEntries{} entries, the geometric-mean kernel speedup is \NPUGeomeanAnyValidSpeedup$\times$ and the median operator speedup is \NPUMedianAnyValidSpeedup$\times$.
The level-wise geometric means in Table~\ref{tab:npukernelbench-summary} differ substantially.
Figure~\ref{fig:npukernelbench-runtime} reports the associated geometric-mean kernel times before and after optimization.
This level difference is descriptive rather than causal: the two benchmark groups contain different operators and shapes, so it does not isolate a benefit of any component of \system{}.

\subsection{RQ2: Outcome Distribution}

Figure~\ref{fig:npukernelbench-speedup} shows a highly heterogeneous distribution.
\NPUAboveTwoXAnyValidEntries{} of \NPUAnyValidEntries{} entries exceed 2.0$\times$, and \NPUAboveFiveXAnyValidEntries{} exceed 5.0$\times$.
The gap between the \NPUGeomeanAnyValidSpeedup$\times$ geometric mean and the \NPUMedianAnyValidSpeedup$\times$ median reflects this long-tailed behavior: a set of large gains coexists with near-baseline entries.

This distribution identifies both opportunity and boundary.
The workflow can produce substantial wins, but the current experiment does not establish uniform improvement or causally isolate the contribution of its four evidence levels.
The remaining analyses therefore ask a narrower, auditable question: how often did optimization need to progress beyond its shallowest evidence level?

\subsection{Comparison with Torch NPU}

The Torch NPU comparison is a separate post-optimization evaluation, not a substitute for the Triton-before/after result above.
For each operator with a matching report, we compare the optimized Triton implementation with its Torch NPU counterpart over all available benchmark cases.
The remaining \NPUTorchTotalEntries{} operators cover \NPUTorchTotalValidCases{} cases (\NPUTorchMinimumCasesPerEntry{}--\NPUTorchMaximumCasesPerEntry{} per operator; median \NPUTorchMedianCasesPerEntry{}).

Table~\ref{tab:torch-npu-comparison} reports a \NPUTorchGeomeanSpeedup$\times$ geometric-mean speedup and a \NPUTorchMedianSpeedup$\times$ median, where speedup is Torch NPU total time divided by optimized Triton total operator time.
\NPUTorchImprovingEntries{} of \NPUTorchTotalEntries{} entries are non-regressing, while \NPUTorchAboveTwoXEntries{} exceed 2$\times$ and \NPUTorchAboveFiveXEntries{} exceed 5$\times$.
Because this protocol uses different case coverage and a different baseline, it provides an independent post-optimization comparison rather than the Triton-before/after optimization result.

\NPUTorchComparisonTable

\subsection{RQ3: Evidence Accounting and Mechanism}

The per-round optimization trajectories let us separate final outcome measurement from process accounting.
For each of the \NPURoundAccountingEntries{} entries, we consider only rounds that pass both correctness and benchmark validation, compare their geometric-mean speedups under that round's declared authority metric, and select the highest-speedup round (breaking ties by the earlier round).
We assign the selected round to its primary evidence level: compiler-source, IR, and profiling evidence take precedence when they are attached to the round; otherwise, its explicitly stated pattern-triage label is used.
Panel~(b) therefore labels the evidence level of the selected best round, whereas panel~(c) counts whether the complete trajectory reaches each level at least once.
The best result is often not immediate: \NPUBestRoundNineToFifteenCount{} of \NPURoundAccountingEntries{} entries first reach their best validated round after round~8, including \NPUBestRoundThirteenToFifteenCount{} in the final three rounds; the median best-round index is \NPUMedianBestRound{}.
This pattern does not prove that every late round contributes a new mechanism, but it shows that later rounds remain empirically active rather than being a purely ceremonial stopping period.

Under this accounting, \NPUWinningLOneCount{} selected best rounds are attributed to pattern triage, \NPUWinningLTwoCount{} to profiling diagnosis, and none to IR attribution or compiler-source escalation.
This does not imply that deep analysis is unused or ineffective: \NPUPathUsesLTwoCount{} trajectories reach profiling diagnosis, \NPUPathUsesLThreeCount{} reach IR attribution, and \NPUPathUsesLFourCount{} reach compiler-source escalation at least once.
The distinction reflects the hierarchy's intended role.  IR and compiler-source analysis can explain a plateau, reject an incompatible pattern, or establish a backend limitation; a subsequent retained candidate can still be a simpler pattern-level rewrite.
The accounting does not establish that deep analysis caused that rewrite to be selected, nor that it generally discovers the final best pattern.
Figure~\ref{fig:npu-optimization-accounting} reports these counts.
This is process evidence, not an ablation: it establishes that the hierarchy is exercised in practice, while a controlled comparison against profiling-only or always-deep policies remains future work.

\paragraph{Representative cross-layer trace.}
The \textsc{Hyena} FFT-size padding operator is a representative trace because it combines profiler, IR, and compiler-source evidence in a single, validated optimization trajectory.
The initial rewrite replaces a flat one-dimensional index decoder with two-dimensional row--column tiling, eliminating per-element division and remainder operations; it reaches 1.89$\times$ on the optimization benchmark.
The agent then profiles the remaining kernel bottleneck and records:

\paperanalysisnote{Agent profiling record.}{``\texttt{aiv\_scalar\_ratio=48.1\%, aiv\_vec\_ratio=12.5\%}.''}

This directs a narrow semantic simplification rather than a blind launch-configuration search: because masked loads already supply zeros outside the live extent, the agent removes a redundant \texttt{tl.where}.  That candidate reaches 2.08$\times$ geometric-mean kernel speedup on the per-round benchmark.
The retained two-dimensional tiled kernel, shown in Listing~\ref{lst:hyena-tiled-rewrite}, combines the removal of the original per-element division and remainder operations with this zero-preserving simplification.

When later source rewrites fail to improve this candidate, the agent escalates to IR attribution.

\paperanalysisnote{Agent IR record.}{``\texttt{hivmave-scalar-broadcast-to-vload} pass adds 2 loads, 20 sync ops.''}

The compiler-source inspection then identifies the corresponding lowering mechanism.

\paperanalysisnote{Agent compiler-source record.}{``Boolean 2D broadcast masks trigger \texttt{brc\_scalar\_core\_1d}.''}

These findings explain why plausible alternatives---static row unrolling, split loops, a one-dimensional grid, and mask-free fast paths---all regress: they either increase program/branch overhead or add memory traffic without removing the dominant lowering cost.
The agent consequently retains the simpler two-dimensional tiled kernel with the redundant conditional removed, rather than claiming that deeper analysis itself produced a new winning rewrite.
Thus the trace shows the hierarchy's practical value as diagnosis and rejection: profiling identifies the symptom, IR localizes its generated cost, compiler-source analysis explains the backend constraint, and validation selects the robust implementation, which reaches 2.08$\times$ on the per-round optimization benchmark.

\begin{papercodeblock}
\begin{lstlisting}[style=diffstyle,caption={Source-level rewrite for the \textsc{Hyena} FFT-size padding operator.},label={lst:hyena-tiled-rewrite}]
 def pad_scale(...):
-    offsets = pid * BLOCK_SIZE + tl.arange(0, BLOCK_SIZE)
-    row, col = offsets // fft_size, offsets % fft_size
-    x = tl.load(x_ptr + row * seqlen + col, mask=load_mask, other=0.0)
-    tl.store(out_ptr + offsets, tl.where(col < seqlen, x * scale, 0.0),
-             mask=store_mask)
+    rows = pid * BLOCK_ROWS + tl.arange(0, BLOCK_ROWS)
+    for col0 in tl.range(0, fft_size, BLOCK_COLS):
+        cols = col0 + tl.arange(0, BLOCK_COLS)
+        src = (rows * seqlen)[:, None] + cols[None, :]
+        dst = (rows * fft_size)[:, None] + cols[None, :]
+        x = tl.load(x_ptr + src, mask=load_mask, other=0.0)
+        tl.store(out_ptr + dst, x * scale, mask=store_mask)
\end{lstlisting}
\end{papercodeblock}

\section{Related Work}

\paragraph{Compiler substrates and compiler-centric tooling.}
Triton provides the immediate DSL and compilation substrate for our setting, while Triton-Ascend extends that stack to Ascend NPUs by adapting the backend to a different lowering pipeline~\cite{tillet2019triton,tritonascendrepo}.
Classical tensor compilers and autotuning systems, exemplified by TVM, cast operator optimization as schedule or configuration search inside a compiler-controlled transformation space~\cite{chen2018tvm}.
This line established the importance of hardware-aware search and compiler-managed optimization.
It does not, however, address an LLM agent that edits an existing Triton kernel while trying to explain the observed slowdown through compiler evidence.
KPerfIR is the closest conceptual precedent on the tooling side: it integrates profiling into a compiler workflow so that performance analysis can be expressed through IR-level abstractions and compiler passes~\cite{guan2025kperfir}.
Our work shares the compiler-centric intuition, but uses that evidence to steer an agentic rewrite loop rather than to introduce a new profiling infrastructure.

\paragraph{LLM-driven kernel synthesis and source-level optimization.}
Recent LLM-based systems show that kernel quality can improve substantially when execution feedback is turned into an optimization signal.
SparseRL~\cite{wang2026sparserl}, Kevin~\cite{baronio2025kevin}, CUDA-L1~\cite{li2025cudal1}, and KernelEvolve~\cite{liao2025kernelevolve} demonstrate that source-level reward loops can raise correctness and throughput for CUDA kernels at scale.
The common pattern is clear: the optimizer mutates kernel text and uses runtime reward, compilation success, or both to decide the next step.
These systems are strong comparators because they move beyond one-shot generation, but their evidence is still dominated by execution outcomes rather than explicit compiler-level attribution.
Benchmarks such as KernelBench, TritonBench, and MultiKernelBench provide the evaluation substrate for this line of work by making correctness and performance measurable~\cite{ouyang2025kernelbench,li2025tritonbench,wen2025multikernelbench}.

\paragraph{Profiling-guided and agentic optimization.}
A second line of work gives the optimizer stronger execution-side or agentic control.
AVO is especially relevant because it elevates a coding agent from candidate generator to autonomous variation operator for GPU attention kernels, with long-horizon edit, test, and revision loops over expert-level implementations~\cite{chen2026avo}.
Its core contribution is therefore agentic search autonomy rather than compiler-level attribution.
TritonForge, PRAGMA, Astra, and KForge use profiling, execution traces, and multi-agent coordination to steer iterative optimization~\cite{li2025tritonforge,lei2025pragma,wei2025astra,sereda2025kforge}.
These systems improve over pure reward-only search because they can localize bottlenecks and preserve good candidates across rounds.
Even so, their evidence remains anchored primarily at the execution level: profiling can identify a hotspot, but it does not by itself explain which lowering decision, legality condition, or pass transition made the hotspot appear in the first place.
Our work is therefore complementary but stricter in scope: it asks the agent to move from profiling to IR attribution and then to compiler-grounded explanation before rewriting source.

\paragraph{NPU-specific kernel generation.}
Very recent NPU-focused systems such as AscendKernelGen and AscendCraft show that vendor-specific DSLs, limited public exemplars, and backend constraints make accelerator-specific kernel generation materially harder than the GPU setting~\cite{cao2026ascendkernelgen,wen2026ascendcraft}.
These works are highly relevant because they validate LLM-based generation on Ascend, but they remain generation-centric.
In contrast, our system starts from existing Triton-Ascend kernels, applies pattern triage when possible, and escalates through profiling diagnosis, IR attribution, and compiler-source escalation before proposing evidence-backed rewrites.
That difference matters: our goal is not only to synthesize a new kernel, but to explain and repair a kernel whose bottleneck is already visible at runtime.

\section{Discussion}

\system{} provides evidence that compiler-grounded, hierarchical optimization can unlock substantial gains, but the latest results also clarify its current boundary.
On the NPUKernelBench-derived Ascend 950 evaluation, the workflow reaches a geometric-mean kernel speedup of \NPUGeomeanAnyValidSpeedup$\times$ and a median of \NPUMedianAnyValidSpeedup$\times$ across \NPUAnyValidEntries{} entries.
The current result remains heterogeneous: large wins coexist with near-baseline entries.

The optimization accounting also shows why the workflow should not be summarized only by its final speedup table.
The median best-round index is \NPUMedianBestRound{}, and \NPUBestRoundThirteenToFifteenCount{} selected paths first peak in the final three rounds.
Moreover, \NPUPathUsesLTwoCount{} paths reach profiling diagnosis, \NPUPathUsesLThreeCount{} reach IR attribution, and \NPUPathUsesLFourCount{} reach compiler-source escalation.
These counts show practical use of progressive evidence, but they do not make a causal claim that any one level is responsible for the final outcome.

What the evaluation does not yet provide is causal isolation.
It does not compare progressive escalation against profiling-only optimization, always-on deep analysis, or other control strategies under one frozen toolchain.
It also does not separate the effect of the retrospective pattern-synthesis loop from the seeded pattern guidance that initializes pattern triage, nor does it report token cost or wall-clock optimization time.
Future work should separately report the optimization cost and marginal value of different round budgets.

The current implementation choices also define the system's boundary.
Restricting rewrites to structured source transformations improves auditability and reduces invalid edits, but it may exclude optimizations that require larger refactors or multi-kernel coordination.
More broadly, transfer to other backends will depend on how much useful compiler-grounded evidence their profiler, IR stack, and source references can expose.
The next step is therefore a stronger systems evaluation with frozen-toolchain baselines, explicit cost accounting, and direct ablations of both the hierarchy and the retrospective synthesis loop.

\section{Conclusion}

We presented a hierarchical and compiler-grounded framework for LLM-based kernel optimization.
Its central idea is to treat optimization as a progressive cross-layer diagnosis problem rather than a one-shot source rewriting task, escalating from pattern triage and profiling to IR attribution and compiler-source escalation before proposing evidence-backed rewrites.
On the current NPUKernelBench-derived Ascend 950 evaluation, \system{} reaches a geometric-mean initial-to-optimized Triton kernel speedup of \NPUGeomeanAnyValidSpeedup$\times$ and a median of \NPUMedianAnyValidSpeedup$\times$ across \NPUAnyValidEntries{} entries.
The full distribution ranges from near-baseline entries to large wins, while the optimization histories show that many selected results peak late and that multiple evidence depths are exercised before convergence.
The current scope remains limited by the lack of causal ablations, frozen-toolchain comparisons against alternative reasoning strategies, and explicit optimization-cost reporting; these form the next step for evaluating the hierarchy.

\balance
\begingroup
\raggedright
\bibliographystyle{plain}
\bibliography{refs}
\endgroup

\end{document}